\documentclass{article}
\usepackage{ijcai26}
\usepackage{times}
\usepackage{soul}
\usepackage{url}
\usepackage[hidelinks]{hyperref}
\usepackage[utf8]{inputenc}
\usepackage[small]{caption}
\usepackage{graphicx}
\usepackage{amsmath}
\usepackage{amsthm}
\usepackage{booktabs}
\usepackage{algorithm}
\usepackage{algorithmic}
\usepackage[switch]{lineno}

\title{A Multimodal Approach to Alzheimer’s Diagnosis: Geometric Insights from Cube Copying and Cognitive Assessments}
\author{
Jaeho Yang$^1$
\and
Kijung Yoon$^{1,2}$
\\
\affiliations
$^1$Department of Electronic Engineering, Hanyang University, Seoul, Korea 04763\\
$^2$Department of Artificial Intelligence, Hanyang University, Seoul, Korea 04763\\
\emails
\{jaehoyang, kiyoon\}@hanyang.ac.kr}

\begin{document}
\maketitle
\begin{abstract}
    Early and accessible detection of Alzheimer’s disease (AD) remains a critical clinical challenge, and cube-copying tasks offer a simple yet informative assessment of visuospatial function. This work proposes a multimodal framework that converts hand-drawn cube sketches into graph-structured representations capturing geometric and topological properties, and integrates these features with demographic information and neuropsychological test (NPT) scores for AD classification. Cube drawings are modeled as graphs with node features encoding spatial coordinates, local graphlet-based topology, and angular geometry, which are processed using graph neural networks and fused with age, education, and NPT features in a late-fusion model. Experimental results show that graph-based representations provide a strong unimodal baseline and substantially outperform pixel-based convolutional models, while multimodal integration further improves balanced classification performance and discriminative ability. SHAP-based interpretability analysis identifies specific graphlet motifs associated with corner integrity and edge continuity as key predictors, closely aligning with clinical observations of distorted cube drawings in AD. Together, these findings establish graph-based analysis of cube-copying behavior as an interpretable, non-invasive, and scalable framework for Alzheimer’s disease screening.
\end{abstract}

\section{Introduction}\label{sec:intro}
Alzheimer's disease (AD) is a progressive neurodegenerative disorder and the most common cause of dementia worldwide \cite{winblad2016defeating,alzheimers2020facts}. It leads to memory loss, cognitive decline, and functional impairment, profoundly affecting patients and their caregivers. Early diagnosis of AD is critically important since it allows patients and families to plan ahead and access treatments or interventions that may slow symptom progression and improve quality of life \cite{jack2018nia}. Traditionally, AD diagnosis has relied on clinical evaluations such as patient history and neurological exams, standardized cognitive tests, and neuroimaging or biomarker analyses \cite{dubois2007research,jack2010hypothetical}. However, these conventional approaches have notable limitations. They often detect the disease only at a relatively late stage and can be subjective or resource intensive. Although non-invasive neuroimaging techniques such as MRI are commonly used in AD assessment, these methods can still be resource-intensive and less accessible in some clinical and community contexts \cite{winblad2016defeating}.

To address these challenges, researchers have explored simpler neuropsychological tests and technology-based solutions that make AD screening faster and easier. Comprehensive neuropsychological assessments, while informative, are time consuming and impractical for routine use \cite{snyder2014developing}. This is especially problematic given the rising number of patients and the limited availability of specialists \cite{prince2013global}. One promising alternative is the cube copying test, a brief visuospatial drawing task in which an individual is asked to copy a wire-frame drawing of a three-dimensional cube \cite{shimada2006necker}. This task is very simple and language independent, and can be performed almost anywhere, even on electronic devices, making it an attractive and portable screening tool. Prior studies have investigated the feasibility of cube copying for dementia detection. Results show that cube-copying performance reflects visuospatial and constructional function in AD and is associated with disease severity \cite{palmqvist2008usefulness}. Patients with AD often produce characteristic errors when copying a cube such as misaligned connections or distorted angles due to their visuospatial deficits \cite{shimada2006necker}. Furthermore, longitudinal research on individuals with mild cognitive impairment (MCI) has shown that those with poor cube-copying scores are significantly more likely to progress to AD, highlighting the test’s potential value for early diagnosis \cite{buchhave2008cube}. These findings suggest that the cube copying task can serve as a sensitive indicator of the subtle cognitive declines associated with AD.

Instead of relying on manual scoring or conventional statistical evaluations of cube drawings \cite{kipf2017semi}, we propose a deep learning approach to uncover geometric biomarkers of AD from the cube copying task. Specifically, we employ graph neural networks (GNNs), a class of neural network architectures designed to operate on graph-structured data \cite{freedman1994clock}. In our framework, each hand-drawn cube image is converted into a graph representation by extracting the key line segments and their intersections. This vectorization is achieved using line-simplification algorithms such as the Ramer Douglas Peucker method \cite{douglas1973algorithms}, which reduce the drawing to a set of nodes representing cube corner points connected by edges representing drawn lines. The resulting graph captures the essential geometry of the sketch and is fed into a GNN model that learns latent representations of the drawing’s shape. We train the GNN to classify subjects into categories of cognitive impairment, including cognitively normal (CN) or AD, based on cube-graph features. Furthermore, we integrate additional modalities including neuropsychological test scores and demographic information such as age and years of education within a multimodal deep learning framework. By fusing drawing-based graph features with clinical and demographic features, we evaluate how a combined multimodal approach improves diagnostic performance. This strategy echoes recent research showing that deep learning models benefit from incorporating diverse data types for more accurate AD classification \cite{jo2019deep,zhang2011multimodal}. Ultimately, our goal is to determine whether the way a patient draws a simple cube can be a reliable determinant for AD diagnosis and to identify which geometric drawing features are most indicative of cognitive impairment.

\section{Related Work}
\paragraph{Neuropsychological Drawing Tasks for Cognitive Assessment.}
Traditional clinical studies established cube copying as a fast visuoconstructional test correlated with global cognition \cite{maeshima2004usefulness,mori2014clinical}. Its sensitivity to early Alzheimer’s disease (AD) has been reported in mild cases where other screening tools are less effective \cite{shimada2006necker}. Longitudinal analyses have further shown that cube-copying scores can track disease progression and treatment response, sometimes capturing preserved visuospatial ability even when global metrics such as the mini-mental state examination (MMSE) decline \cite{palmqvist2008usefulness}. Methodological extensions have considered process-level features such as drawing time, hesitation, and stroke sequence, highlighting the diagnostic value of temporal patterns beyond static outcomes \cite{satoh2016improved}. While these studies support the clinical utility of cube copying, they rely primarily on manual scoring and conventional psychometric analysis, limiting scalability and interpretability.

\paragraph{Automated Analysis via Machine Learning.}
The digitization of drawing tasks such as cube copying, clock drawing (CDT), and trail-making has enabled algorithmic analysis. Early machine learning approaches engineered behavioral features—including stroke timing, pen pressure, and error patterns—and applied classifiers to distinguish cognitively impaired individuals from controls \cite{bennasar2014cascade,souillard2016learning,davis2014think}. These methods demonstrated that low-level motor and process features encode diagnostic signals, but they were constrained by manual feature design and modest generalizability.

\paragraph{Deep Learning for Drawing-Based Screening.}
Recent advances leverage convolutional neural networks (CNNs) and related architectures to process raw drawing images or temporal traces directly. CNN-based pipelines have achieved strong performance in CDT image classification \cite{chen2020automatic,park2021automatic}, while online systems have explored AI-assisted screening in clinical and mobile environments \cite{amini2021ai}. Multimodal extensions, such as integrating CDT with Rey-Osterrieth figure copying, have shown that fusing heterogeneous visuospatial tasks yields more robust classification \cite{youn2021use}. Nevertheless, CNN-based models primarily exploit pixel-level features and lack explicit representations of underlying geometric structures, which are central to visuoconstructional tasks.

\paragraph{Multimodal and Explainable AI Approaches.}
More recent efforts emphasize multimodality and interpretability. For instance, explainable deep models combining CDT, cube-copying, and trail-making tasks have employed attention mechanisms to generate clinically relevant saliency maps \cite{ruengchaijatuporn2022explainable}. Such approaches demonstrate the potential of combining drawing data with other modalities to enhance both accuracy and clinical trust. However, most prior work focuses on CDT and relies on grid- or image-based representations, without fully exploiting the inherent graph structure of geometric drawings such as the cube.

\paragraph{Positioning Our Approach.}
Our work bridges these gaps by introducing GNNs to analyze cube drawings as graph-structured objects. Unlike CNN-based pixel approaches or hand-crafted feature pipelines, our method vectorizes sketches into nodes and edges, preserving geometric fidelity while enabling principled representation learning. We further embed these graph-derived features in a multimodal framework alongside neuropsychological scores and demographic variables. This design both leverages the geometric essence of visuoconstructional tasks and aligns with the emerging trend toward multimodal, interpretable AI for dementia diagnosis.

\begin{figure*}[t]
    \centering
    \includegraphics[width=0.9\linewidth]{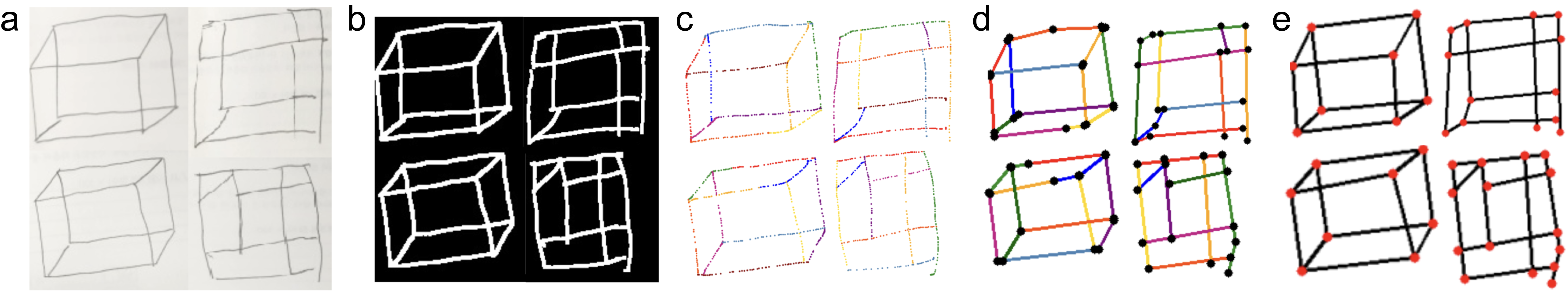}
    \caption{Pipeline for transforming hand-drawn cube images into graph-structured representations. (a) Raw cube drawings collected from participants. (b) Binarized images obtained by adaptive thresholding and noise filtering. (c) Vectorized line representations produced from skeletonized strokes. (d) Line simplification and node extraction, showing intermediate polylines with candidate junctions. (e) Final graph representations after merging nearby nodes and pruning spurious ones, where nodes correspond to cube corners and edges correspond to drawn line segments.}
      \label{fig:pipeline}
\end{figure*}

\section{Methods}
We propose a multimodal framework for Alzheimer’s disease classification that integrates graph-based representations of hand-drawn cubes with demographic and neuropsychological test (NPT) information. In this study, the NPT corresponds to the Seoul Neuropsychological Screening Battery (SNSB), a standardized cognitive assessment battery that evaluates multiple cognitive domains, including memory, attention, language, visuospatial ability, and executive function. Each domain is quantitatively scored based on task performance, and the resulting scores provide a comprehensive measure of an individual's cognitive function. The core idea is to model each hand-drawn cube as a graph that preserves its geometric and structural characteristics, and to fuse this graph-based representation with complementary clinical features within a unified predictive model.
This section details the proposed methodology, including the graph construction pipeline (\ref{method:graph_pipeline}), node feature representation (\ref{method:node_features}), graph neural network architecture (\ref{method:gnn}), and multimodal integration strategy (\ref{method:multimodal}). We also describe the dataset (\ref{method:dataset}) and evaluation metrics (\ref{method:metrics}), followed by the SHAP-based interpretability analysis.

\subsection{Transforming Hand-Drawn Cube Images into Graph-Structured Data}
\label{method:graph_pipeline}
Our framework begins with the conversion of raw hand-drawn cube sketches (Figure \ref{fig:pipeline}a) into graph representations that capture the geometric structure of the drawing. This pipeline comprises four main steps (Algorithm \ref{alg:graph_pipeline}): adaptive thresholding and noise filtering, vectorization, line simplification, and node refinement.

Given that sketches often contain variable stroke intensity, inconsistent shading, or unintended pen marks, we employ adaptive thresholding to binarize the input image. This ensures that line boundaries accurately represent the drawn cube while suppressing illumination artifacts. Noise filtering is then applied to eliminate extraneous pixels, such as accidental dots or light pen pressure, resulting in a clean black-white image where white pixels represent meaningful strokes and the background remains black (Figure \ref{fig:pipeline}b). Specifically, noise filtering removes isolated connected components below a minimum pixel area threshold, thereby suppressing small artifacts while preserving continuous stroke structures.

Following binarization, we convert rasterized strokes into vectorized representations. As a preprocessing step, we first apply a skeletonization operation that iteratively thins binary strokes to a one-pixel-wide medial axis representation while preserving connectivity and topology. The resulting skeleton image retains the geometric structure of the original drawing while eliminating stroke thickness variability. We then adopt an algorithmic framework for vectorization \cite{bessmeltsev2019vectorization}, which reconstructs curves from skeletonized pixel data by approximating strokes with polylines. This step transforms continuous line segments into explicit coordinate sequences (Figure 1c), thereby facilitating downstream geometric simplification and graph extraction. The output consists of a collection of polylines corresponding to individual edges of the cube sketch.

\begin{algorithm}[t!]
    \caption{Algorithm for transforming hand-drawn cube images into graph-structured data.}
    \label{alg:graph_pipeline}
    \textbf{Input}: Hand-drawn cube image $I$ (raster)\\
    \textbf{Parameter}: Adaptive thresholding method $\mathcal{T}$; noise filter $\mathcal{F}$; vectorization method $\mathcal{V}$; Douglas--Peucker tolerance $\varepsilon = 10$; node merge radius $\delta = 15$\\
    \textbf{Output}: Graph $G=(V,E)$ that models the cube sketch as nodes and edges
    \begin{algorithmic}[1]
        \STATE \textbf{// Step 1: Adaptive thresholding and noise filtering}
        \STATE $B \leftarrow \mathcal{T}(I)$ \COMMENT{Binarize image to suppress illumination artifacts}
        \STATE $B \leftarrow \mathcal{F}(B)$ \COMMENT{Remove spurious pixels (dots, light marks), keep strokes}

        \STATE \textbf{// Step 2: Vectorization}
        \STATE $S \leftarrow \textsc{Skeletonize}(B)$
        \STATE $\mathcal{P} \leftarrow \mathcal{V}(S)$ \COMMENT{Reconstruct strokes as a set of polylines}

        \STATE \textbf{// Step 3: Line simplification}
        \STATE $\mathcal{P}_{simp} \leftarrow \emptyset$
        \FORALL{$P \in \mathcal{P}$}
            \STATE $\hat{P} \leftarrow \textsc{DouglasPeucker}(P,\varepsilon)$
            \STATE $\mathcal{P}_{simp} \leftarrow \mathcal{P}_{simp} \cup \{\hat{P}\}$
        \ENDFOR

        \STATE \textbf{// Step 4: Graph construction and node refinement}
        \STATE $V_{cand} \leftarrow \emptyset$, $E \leftarrow \emptyset$
        \FORALL{$\hat{P} \in \mathcal{P}_{simp}$}
            \STATE $u \leftarrow \textsc{StartPoint}(\hat{P})$, $v \leftarrow \textsc{EndPoint}(\hat{P})$
            \STATE $V_{cand} \leftarrow V_{cand} \cup \{u,v\}$
            \STATE $E \leftarrow E \cup \{(u,v)\}$ \COMMENT{Create an edge per simplified stroke}
        \ENDFOR

        \STATE $V \leftarrow \textsc{MergeNearbyNodes}(V_{cand}, \delta)$ \COMMENT{Cluster nodes within radius $\delta$}
        \STATE $E \leftarrow \textsc{ReindexEdges}(E, V)$ \COMMENT{Replace endpoints by merged canonical nodes}
        \STATE $G \leftarrow (V,E)$
        \STATE \textbf{return} $G$
    \end{algorithmic}
\end{algorithm}

To reduce redundancy in vectorized line data, we employ the Douglas-Peucker algorithm \cite{douglas1973algorithms}, which iteratively removes points lying within a specified tolerance of the original curve. This procedure preserves the overall shape of each line segment while discarding minor deviations caused by noise or hand tremors. By compressing polylines into a smaller set of critical points, we obtain a compact yet accurate representation of the cube’s edges, suitable for graph construction (Figure \ref{fig:pipeline}d).

Finally, we construct a graph by treating polyline endpoints as candidate nodes and line segments as edges. To ensure geometric consistency, nearby nodes are merged into single canonical nodes when they lie within a predefined spatial threshold. The resulting graph consists of well-defined corner nodes connected by edges corresponding to drawn cube lines (Figure \ref{fig:pipeline}e), faithfully representing the intended geometric structure while minimizing noise.

The proposed pipeline contains several tunable parameters, including the noise filtering threshold, Douglas-Peucker tolerance $\varepsilon$, and node merge radius $\delta$. Rather than relying on extensive parameter fine-tuning, these parameters are selected to reflect geometric properties of hand-drawn sketches and are applied consistently across all samples. In practice, the pipeline exhibited stable behavior across a range of parameter values because each stage primarily acts as a coarse structural refinement rather than enforcing strict geometric assumptions. For example, Douglas-Peucker simplification removes only small local perturbations while preserving major line trajectories, and node merging operates only on spatially proximal endpoints. Consequently, moderate variations in parameter values resulted in similar graph structures, suggesting that graph extraction is driven predominantly by the global cube geometry rather than precise threshold selection.

\subsection{Node Feature Representation}\label{method:node_features}
\begin{figure}[t]
    \centering
    \includegraphics[width=0.9\columnwidth]{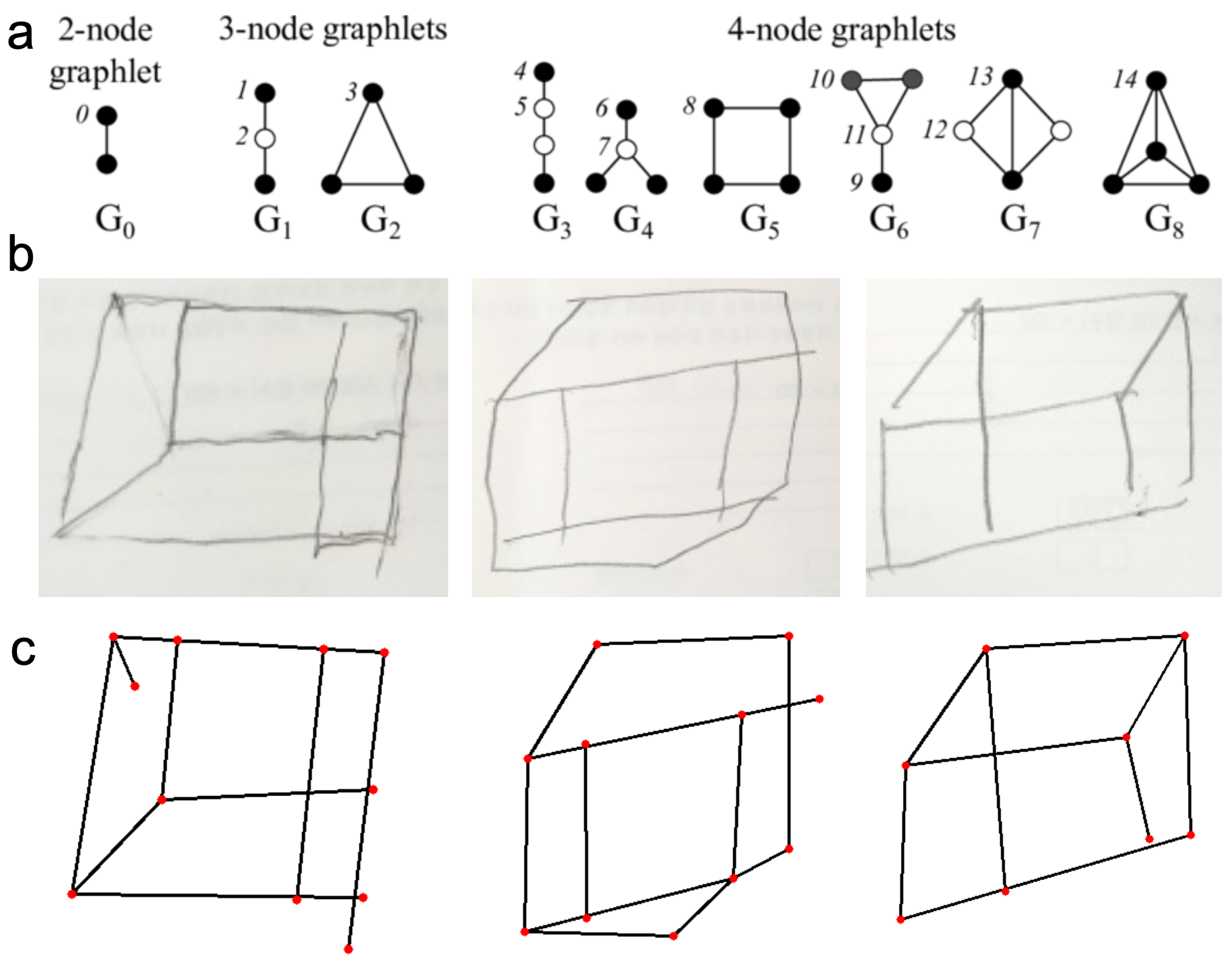}
    \caption{(a) The first 15 graphlets, including 2-, 3-, and 4-node induced subgraphs, with node indices indicating graphlet orbits that specify the structural role of each node within an isomorphism class. (b) Three representative cube drawings produced by AD participants. (c) The corresponding cube graph representations extracted from the drawings, where vertices (red dots) denote junctions or endpoints and edges represent connecting strokes. In panel (a), black and white nodes distinguish different orbit roles within each graphlet, highlighting structurally distinct node positions that collectively characterize the topological signature of the drawing graphs.}
    \label{fig:cube_samples_ad}
\end{figure}

Once cube drawings are transformed into graph-structured data, each node in the resulting graph must be described with features that capture both its geometric placement and its local structural context. We employ three complementary types of node features. First, the absolute (x,y) coordinates of each node are included to preserve the spatial layout of the cube drawing. These coordinates provide direct information about the position of corners in the sketch and enable the model to distinguish between geometrically distinct configurations. Second, we incorporate the graphlet degree vector (GDV) \cite{przulj2007biological}, a structural descriptor based on graphlet counting. A graphlet is a small induced subgraph, and the GDV characterizes how often a node participates in each possible orbit of graphlets \cite{przulj2007biological,przulj2004modeling,sarajlic2016graphlet}. In this study, we use graphlets up to size four, resulting in a 15-dimensional feature vector per node (Figure \ref{fig:cube_samples_ad}a). This representation encodes local topological patterns such as whether a node tends to form part of triangles, squares, or other small motifs. By capturing the structural role of each corner within the cube graph, GDVs provide rich relational context that complements raw coordinates.

Finally, we compute the inner angles formed between edges that share the same node. Since each node in a cube drawing can connect to multiple neighboring nodes, we calculate up to three unique inner angles defined by adjacent edge pairs. These angles encode local geometric consistency, for example whether connected edges approximate right angles or become distorted. This information is particularly relevant for identifying visuospatial impairments reflected in skewed or misaligned corners. The resulting node representation thus concatenates spatial, structural, and geometric descriptors, yielding a feature vector that balances positional fidelity, relational structure, and angular geometry for downstream graph neural network modeling.

\subsection{Graph Neural Network Modeling}\label{method:gnn}
The cube drawings represented as graphs, with nodes described by spatial, structural, and geometric features, serve as input to GNNs. Each node is embedded into a latent representation through message passing, where information is propagated along edges to capture both local geometry and global cube structure. In this study, we adopt the graph attention network (GAT) architecture \cite{velickovic2018graph,brody2022how}, which leverages self-attention mechanisms to assign different importance weights to neighboring nodes during message passing. This attention-based aggregation enables the model to focus on the most informative local relationships, improving its ability to capture heterogeneity in node connectivity and structural patterns within the cube graphs. We use the GraphGym library \cite{you2020design} to systematically explore architectural choices including the number of layers, hidden dimensions, aggregation schemes, and normalization strategies. GraphGym enables automated and reproducible benchmarking across a wide design space, allowing us to identify the most effective configuration for distinguishing between control and AD. The output of the GNN model is a graph-level representation that summarizes the geometric and structural features of the cube drawing, which is then forwarded to the multimodal integration stage.

\begin{figure}[t]
    \centering
    \includegraphics[width=0.95\columnwidth]{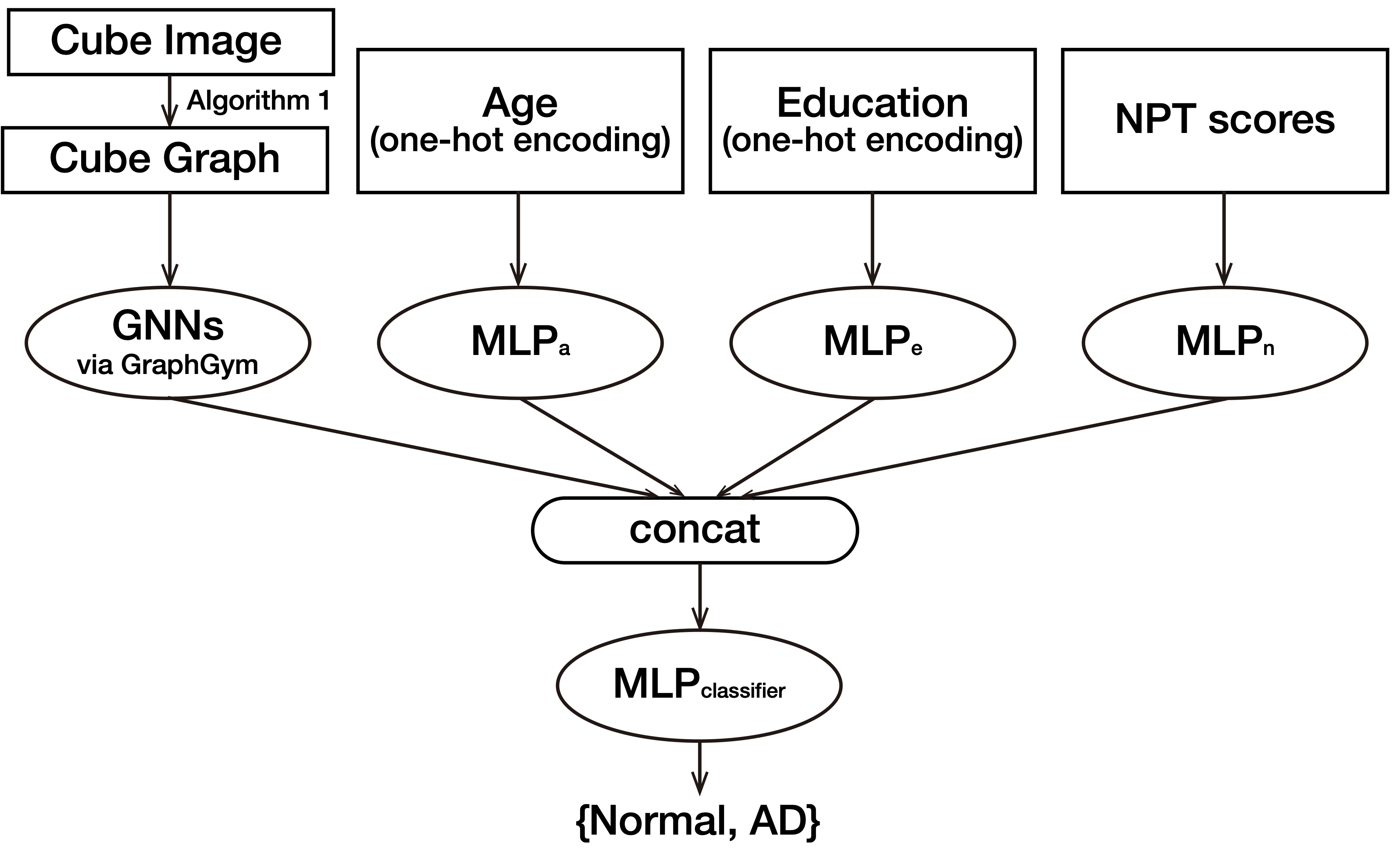}
    \caption{Multimodal classification framework integrating cube drawing graphs and clinical features. The cube image is first converted into a cube graph using Algorithm \ref{alg:graph_pipeline} and processed by a graph neural network GNN implemented via GraphGym to extract a graph-level representation. Demographic variables (age and education, one-hot encoded) and neuropsychological test scores are independently encoded using modality-specific MLPs. The resulting embeddings from all modalities are concatenated and passed to a final MLP classifier to predict diagnostic status (Normal vs. AD).}
    \label{fig:model_framework}
\end{figure}

\subsection{Multimodal Integration with Clinical and Demographic Features}\label{method:multimodal}
In addition to graph-derived features, we incorporate auxiliary modalities including age, years of education, and NPT scores. The age and years of education variables are represented as categorical variables following the original format provided in SNSB. Specifically, age is categorized into nine groups (45-49, 50-54, 55-59, 60-64, 65-69, 70-74, 75-79, 80-84, and 85-90) and encoded as a one-hot vector. Similarly, years of education are categorized into seven groups (illiterate, 0-2 years, 3-5 years, 6-9 years, 10-12 years, 13-15 years, and 16+ years) and represented using one-hot encoding. In contrast, NPT scores are provided on different numerical scales across cognitive tasks and domains. To enable consistent integration across heterogeneous assessments, all NPT scores are rescaled to the range between 0 and 1 before being used as model inputs. Each modality is then passed through a three-layer multilayer perceptron (MLP) to obtain dense embeddings in a shared latent space. These embeddings capture complementary information about demographic and cognitive background that may not be directly reflected in cube drawings.

The outputs from the GNN and the modality-specific MLPs are concatenated into a unified representation. This joint feature vector is processed by a two-layer MLP classifier trained to predict cognitive status (control or AD). The design allows the model to exploit both geometric biomarkers extracted from cube copying and auxiliary clinical information, improving robustness and classification accuracy. By merging heterogeneous modalities in a late-fusion framework, the system integrates structural, demographic, and cognitive perspectives into a single predictive model (Figure \ref{fig:model_framework}).

\begin{table}[t]
\centering
\caption{Demographic summary of participants across cognitive status, age groups, and education levels.}
\resizebox{0.5\linewidth}{!}{
\begin{tabular}{lcc}
\toprule
\textbf{Category} & \textbf{CN} & \textbf{AD} \\
\midrule

\multicolumn{3}{l}{\textit{Cognitive status}}\\
Total participants & 96 & 28 \\

\midrule
\multicolumn{3}{l}{\textit{Age (years)}}\\
45--49 & 0 & 0\\
50--54 & 2 & 2\\
55--59 & 9 & 1\\
60--64 & 15 & 0\\
65--69 & 21 & 2\\
70--74 & 17 & 3\\
75--79 & 25 & 11\\
80--84 & 7 & 9\\
85--89 & 0 & 0\\

\midrule
\multicolumn{3}{l}{\textit{Education (years)}}\\
Illiterate & 7 & 5\\
1--3 & 22 & 9\\
4--6 & 13 & 4\\
7--9 & 24 & 5\\
10--12 & 0 & 0\\
13--15 & 20 & 2\\
16+ & 10 & 3\\

\bottomrule
\end{tabular}}
\label{tab:demographics}
\end{table}

\subsection{Dataset}
\label{method:dataset}

The data used in this study were collected from 124 participants recruited at multiple neurology clinics and community dementia centers over a three-year period, including 28 individuals diagnosed with Alzheimer’s disease. To maximize data utilization while ensuring robust evaluation under the limited sample size, we adopted a stratified 5-fold cross-validation protocol rather than a fixed train-validation-test split. All data partitions were performed at the participant level to prevent data leakage across folds, and class proportions were preserved within each fold through stratified sampling. Given the relatively small number of AD participants, we additionally enforced a minimum of five AD cases per fold to maintain stability and reduce performance variance across cross-validation splits.

Table~\ref{tab:demographics} summarizes the demographic composition of the cohort, including cognitive status, age ranges, and years of education. The dataset consists of 96 cognitively normal participants and 28 individuals with AD. AD participants were more concentrated in older age groups, particularly between 75 and 84 years, whereas CN participants were more broadly distributed across middle-to-late adulthood. Education-level distributions also varied across groups, with AD participants appearing more frequently in lower education categories.

\subsection{Ethical Considerations}
This study was approved by the Institutional Review Board at Hanyang University Medical Center (IRB No. 2023-06-019) and conducted in accordance with the ethical standards outlined in the 1964 Declaration of Helsinki and its later amendments. Written informed consent was obtained from all participants or their legally authorized representatives, as applicable.

\begin{table*}[t]
    \centering
    \caption{Comparison of different multimodal combinations for AD classification, with cube graph alone serving as the unimodal baseline. Results are reported as mean $\pm$ standard error of the mean (SEM) across 10 trials of 5-fold cross-validation. Bold values indicate the best performance for each metric, while underlined values denote the second-best performance. The multimodal model combining cube graphs with age, education, and NPT achieves the strongest overall performance across most evaluation metrics.}
    \resizebox{\textwidth}{!}{%
    \begin{tabular}{lccccccc}
    \toprule
    Model & Accuracy & F1 & AUROC & AUPRC & Precision & Recall & Balanced Acc. \\
    \midrule
    Cube graph + Age + Edu + NPT 
    & \textbf{0.909 $\pm$ 0.010}
    & \textbf{0.750 $\pm$ 0.035}
    & \textbf{0.942 $\pm$ 0.016}
    & \underline{0.876 $\pm$ 0.011}
    & \textbf{0.938 $\pm$ 0.025}
    & \underline{0.656 $\pm$ 0.017}
    & \underline{0.819 $\pm$ 0.012}
    \\

    Cube graph + Age + Edu
    & \underline{0.904 $\pm$ 0.021}
    & \underline{0.746 $\pm$ 0.066}
    & 0.906 $\pm$ 0.029
    & 0.846 $\pm$ 0.045 
    & 0.856 $\pm$ 0.073
    & \textbf{0.692 $\pm$ 0.085}
    & \textbf{0.829 $\pm$ 0.042}
    \\

    Cube graph + Age + NPT
    & 0.881 $\pm$ 0.026 
    & 0.667 $\pm$ 0.084 
    & 0.930 $\pm$ 0.032 
    & 0.868 $\pm$ 0.038 
    & 0.861 $\pm$ 0.072 
    & 0.578 $\pm$ 0.081 
    & 0.774 $\pm$ 0.044
    \\

    Cube graph + Edu + NPT
    & 0.879 $\pm$ 0.024 
    & 0.652 $\pm$ 0.091 
    & 0.918 $\pm$ 0.039 
    & 0.842 $\pm$ 0.050 
    & \underline{0.868 $\pm$ 0.057}
    & 0.562 $\pm$ 0.102 
    & 0.767 $\pm$ 0.051 
    \\

    Cube graph
    & 0.867 $\pm$ 0.032 
    & 0.516 $\pm$ 0.151 
    & \underline{0.939 $\pm$ 0.055}
    & \textbf{0.891 $\pm$ 0.084}
    & 0.794 $\pm$ 0.145 
    & 0.437 $\pm$ 0.154 
    & 0.715 $\pm$ 0.074 
    \\

    Age + Edu + NPT
    & 0.856 $\pm$ 0.020 
    & 0.590 $\pm$ 0.065 
    & 0.879 $\pm$ 0.035 
    & 0.766 $\pm$ 0.047 
    & 0.807 $\pm$ 0.072 
    & 0.498 $\pm$ 0.068 
    & 0.729 $\pm$ 0.036 
    \\
    \bottomrule
    \end{tabular}}
    \label{tab:pred_multimodal}
\end{table*}

\subsection{Evaluation Metrics and Model Interpretation}
\label{method:metrics}
Model selection was based on validation performance, with the optimal model defined as the one achieving the highest macro-F1 score. This criterion is particularly suitable in the presence of class imbalance. For evaluation, we report standard classification metrics including accuracy, F1 score, area under the receiver operating characteristic curve (AUROC), area under the precision-recall curve (AUPRC), precision, recall, and balanced accuracy. Metrics are computed with respect to the positive class, and particular emphasis is placed on AUROC and AUPRC because they provide robust assessments under class imbalance. AUROC reflects the model’s ability to distinguish between AD and non-AD cases across varying decision thresholds, whereas AUPRC is especially informative when positive examples are relatively scarce. Statistical significance of pairwise model comparisons was further evaluated using paired \textit{t}-tests and Kolmogorov–Smirnov (KS) tests, with the resulting \textit{p}-values summarized in Tables~\ref{tab:ttest_pred_multimodal}–\ref{tab:ks_gnn_architecture_comparison} of the Supplementary Material \ref{sec:SI}.

In addition to predictive performance, we employed SHAP (SHapley Additive exPlanations) values \cite{lundberg2017unified} to interpret model outputs and identify features that most strongly contribute to classification decisions. SHAP is based on cooperative game theory and assigns each feature an importance score corresponding to its marginal contribution to the model prediction. The magnitude of a SHAP value reflects the strength of feature influence, while its sign indicates directionality. Positive SHAP values increase the probability of predicting AD, whereas negative values support the non-AD class. By aggregating SHAP values across participants, we obtain a global ranking of influential features that enhances the interpretability of the multimodal model.

\section{Experimental Results}
\subsection{Multimodal Ablation and Performance Analysis}
We evaluate the effect of integrating graph-derived features, demographic variables, and NPT scores. As shown in Table~\ref{tab:pred_multimodal}, the multimodal model combining cube graphs with age, education, and NPT achieves the strongest overall performance, yielding the highest accuracy (0.909), F1 score (0.750), AUROC (0.942), AUPRC (0.876), and precision (0.938). These results suggest that each modality provides complementary information that collectively improves the reliability of AD classification.

The unimodal cube graph model already demonstrates strong discriminative ability, achieving the second-highest AUROC (0.939; $p=0.873$) and AUPRC (0.891; $p=0.562$). However, its lower F1 score (0.516; $p<0.001$), recall (0.437; $p<0.001$), and balanced accuracy (0.715; $p<0.001$) indicate limitations in identifying AD cases consistently under class imbalance. This result suggests that graph representations extracted from hand-drawn cubes capture meaningful structural information but may benefit from additional contextual information. Adding demographic information substantially improves classification performance. In particular, combining cube graphs with age and education increases the F1 score from 0.516 to 0.746 while achieving the highest recall (0.692; $p=0.127$) and balanced accuracy (0.829; $p=0.409$) among all configurations. This finding indicates that demographic variables provide substantial complementary information and improve sensitivity to AD cases. Notably, despite not including NPT information, this configuration performs comparably to the full multimodal model across several metrics.

Including NPT information further improves discriminative performance when integrated with graph and demographic features. The complete multimodal model increases AUROC from 0.906 to 0.942 and AUPRC from 0.846 to 0.876 relative to the Cube graph+Age+Edu configuration, while also improving precision from 0.856 to 0.938. These results suggest that NPT information contributes additional discriminative value, particularly for ranking performance and prediction confidence, even when classification performance is already strong. Similar trends are observed in intermediate combinations involving NPT features. For example, Cube graph+Age+NPT achieves strong AUROC (0.930; $p=0.185$) and AUPRC (0.868; $p=0.418$), while Cube graph+Edu+NPT attains the second-highest precision (0.868; $p<0.001$). These findings suggest that NPT features provide complementary cognitive information but are most effective when integrated with both graph-derived and demographic features.

Finally, using only non-graph modalities (Age+Edu+NPT) results in lower performance across most metrics compared to graph-based multimodal configurations, despite achieving reasonably competitive AUROC (0.879; $p<0.001$). This observation highlights the central role of cube-derived graph representations as a primary source of discriminative information. Taken together, these ablation results suggest that graph features form a strong foundation for AD prediction, while demographic and neuropsychological information provide complementary benefits. The strongest and most balanced performance emerges when all modalities are jointly integrated, demonstrating the value of multimodal fusion for robust AD classification.

\subsection{Interpreting Model Decisions: Global Feature Importance and Structural Roles of Discriminative Graphlets}
\begin{figure}[t]
    \centering
    \includegraphics[width=1.0\columnwidth]{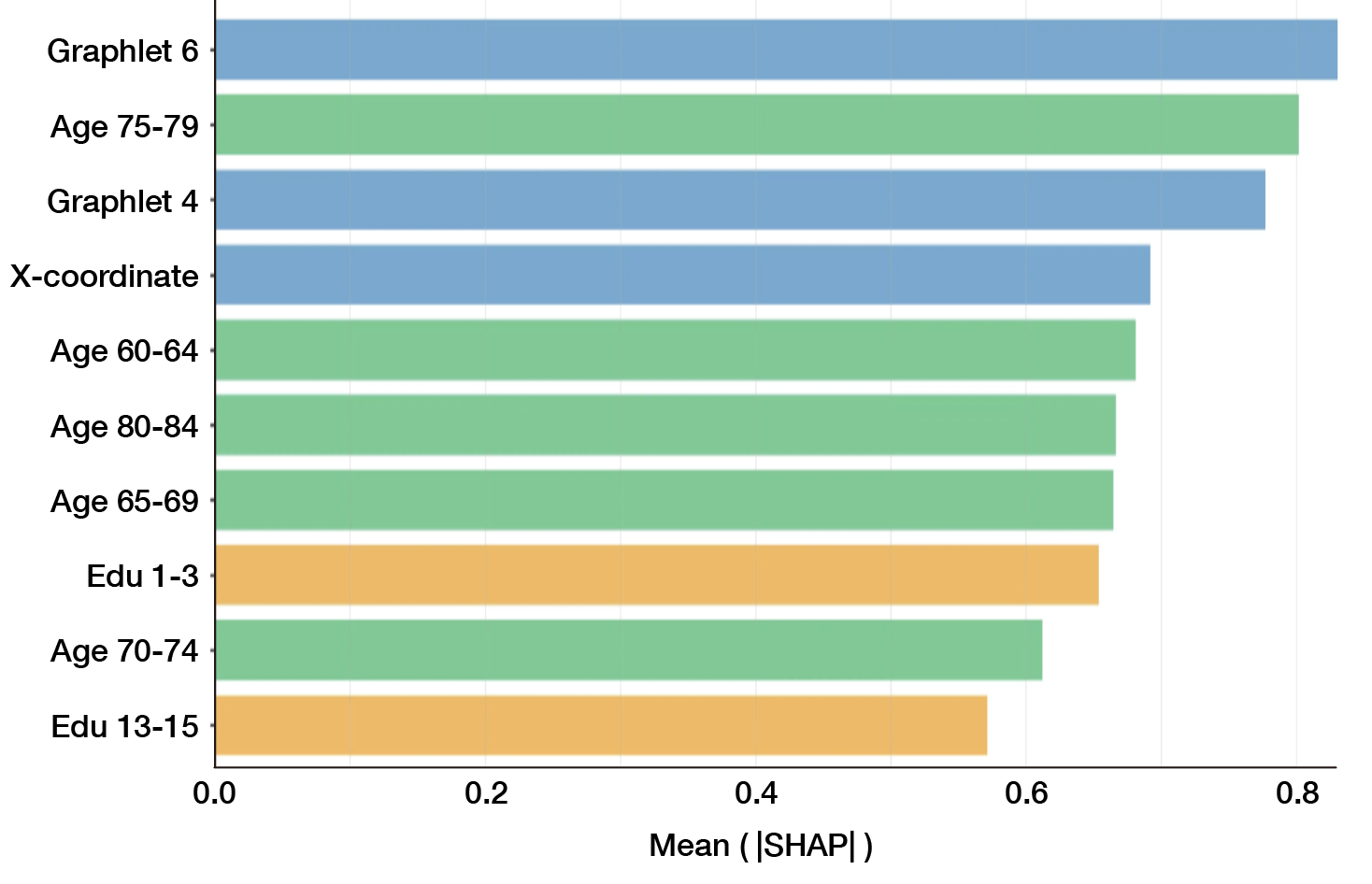}
    \caption{The bar plot shows the top 10 features ranked by mean absolute SHAP values, reflecting their overall contribution to the model’s predictions. Graph-derived features, particularly graphlet 6 and graphlet 4, exhibit the highest importance, highlighting the central role of cube-structural motifs in AD classification. Age-related variables (e.g., 75–79, 60–64, 80–84) also contribute substantially, indicating strong demographic effects, while education levels provide additional complementary information. NPT features are not shown, as they fall outside the top-ranked features.}
    \label{fig:mean_shap}
\end{figure}

To further understand how the model leverages information from each modality, we next examine global feature importance using SHAP analysis. Figure~\ref{fig:mean_shap} presents the global feature importance ranking based on mean absolute SHAP values, summarizing the overall contribution of each feature to the model’s predictions across all samples. Features are grouped by modality, including graph-derived features, demographic variables, and neuropsychological test scores.

Graph-derived features dominate the top of the ranking, with graphlet 6 emerging as the most influential predictor, followed by age (75--79) and graphlet 4. The prominence of graphlet 6 suggests that cube corner structures play an important role in distinguishing AD-related visuospatial impairments, while graphlet 4 similarly suggests the relevance of edge continuity in well-formed cube drawings. The diagnostic relevance of these motifs becomes clearer when considering their expected counts in an ideal cube structure (Figure~\ref{fig:cube_samples_ad}a). Graphlet 6 represents a corner connected to three neighboring vertices and appears three times in a correctly drawn cube. In contrast, graphlet 7 captures a simpler corner structure and occurs only once. Reduced graphlet 6 counts in AD drawings (Figures~\ref{fig:cube_samples_ad}b-\ref{fig:cube_samples_ad}c) therefore suggest disruptions in corner integrity and overall structural fidelity. Graphlet 4 represents a four-node chain motif associated with continuous cube edges and occurs six times in a correctly drawn cube. Distortions, incomplete edge connections, or slight misalignments reduce its frequency, leading to weaker geometric coherence in AD drawings (Figures~\ref{fig:cube_samples_ad}b-\ref{fig:cube_samples_ad}c). Together, the reduced occurrence of graphlet 6 and graphlet 4 motifs in AD samples suggests that visuospatial errors manifest as quantifiable structural disruptions in cube-copying behavior.

Age-related variables also feature prominently in the global importance ranking, with multiple age groups (60--64 through 80--84) appearing among the most influential predictors. This consistent presence indicates that age provides important contextual information that complements the structural information captured by graph-derived features. Education-related variables show a moderate contribution, suggesting a secondary but non-negligible role in shaping classification outcomes. In contrast, NPT features do not appear among the top-ranked variables in Figure~\ref{fig:mean_shap}, indicating that their mean absolute SHAP values are lower than those of the most influential graph and demographic features. Nevertheless, this observation does not imply that NPT information is uninformative. The ablation analysis showed that incorporating NPT features improved AUROC and AUPRC, suggesting that NPT scores provide complementary information whose contribution may arise through interactions with graph-derived and demographic variables rather than strong standalone effects.

Overall, the SHAP analysis reveals a consistent hierarchy of feature importance that closely mirrors the multimodal ablation results. Graph-derived structural features emerge as the primary contributors to prediction, age-related demographics provide substantial complementary information, and NPT variables contribute additional supportive signals through multimodal interactions. Together, these findings suggest that combining structural, demographic, and cognitive information yields both improved predictive performance and enhanced interpretability.

\begin{table*}[t]
    \centering
    \caption{Comparison of different graph-based and pixel-based neural network architectures for AD classification. Results are reported as mean $\pm$ standard error of the mean (SEM) across 10 trials of 5-fold cross-validation. The table contrasts attention-based, geometric, and transformer-style GNNs with conventional CNN and ResNet-18 baselines. For fair comparison, all models were evaluated under the same experimental setting and used identical auxiliary inputs, including age, education level, and NPT information. Bold values indicate the best performance for each metric, while underlined values denote the second-best performance.}
    \resizebox{\textwidth}{!}{%
    \begin{tabular}{lccccccc}
    \toprule
    Model & Accuracy & F1 & AUROC & AUPRC & Precision & Recall & Balanced Acc. \\
    \midrule
    GAT
    & \textbf{0.909 $\pm$ 0.010}
    & \textbf{0.750 $\pm$ 0.035}
    & \textbf{0.942 $\pm$ 0.016}
    & \textbf{0.876 $\pm$ 0.011}
    & \textbf{0.938 $\pm$ 0.025}
    & \textbf{0.656 $\pm$ 0.017}
    & \textbf{0.819 $\pm$ 0.012}
    \\
    \midrule

    Polynormer
    & \underline{0.884 $\pm$ 0.016}
    & \underline{0.684 $\pm$ 0.056}
    & 0.911 $\pm$ 0.023
    & 0.846 $\pm$ 0.037
    & 0.855 $\pm$ 0.095
    & \underline{0.603 $\pm$ 0.059}
    & \underline{0.785 $\pm$ 0.025}
    \\

    EGNN
    & 0.882 $\pm$ 0.025
    & 0.659 $\pm$ 0.077
    & \underline{0.924 $\pm$ 0.037}
    & \underline{0.874 $\pm$ 0.043}
    & \underline{0.899 $\pm$ 0.071}
    & 0.562 $\pm$ 0.085
    & 0.768 $\pm$ 0.044
    \\
    \midrule

    CNN
    & 0.721 $\pm$ 0.048
    & 0.320 $\pm$ 0.055
    & 0.719 $\pm$ 0.043
    & 0.481 $\pm$ 0.044
    & 0.401 $\pm$ 0.121
    & 0.317 $\pm$ 0.036
    & 0.579 $\pm$ 0.042
    \\

    ResNet-18
    & 0.738 $\pm$ 0.030
    & 0.333 $\pm$ 0.092
    & 0.720 $\pm$ 0.046
    & 0.497 $\pm$ 0.061
    & 0.407 $\pm$ 0.132
    & 0.323 $\pm$ 0.094
    & 0.590 $\pm$ 0.050
    \\
    \bottomrule
    \end{tabular}}
    \label{tab:gnn_architecture_comparison}
\end{table*}

\subsection{Effect of GNN Architecture Choice}
Recent advances in graph machine learning have introduced increasingly expressive architectures, including geometric or equivariant GNNs that model rotations and distances \cite{fuchs2020se3,satorras2021en}, as well as transformer-based and motif-aware models designed to capture higher-order subgraph interactions \cite{ying2021do,deng2024polynormer,alsentzer2020subgraph}. While such models offer strong representational capacity, their complexity can obscure interpretability and pose challenges in limited-sample settings.

To contextualize our architectural choice, we evaluated two representative advanced alternatives: EGNN \cite{satorras2021en}, which incorporates geometric equivariance through coordinate-aware message passing, and Polynormer \cite{deng2024polynormer}, a graph transformer capable of modeling long-range dependencies. For a fair comparison, all models in Table~\ref{tab:gnn_architecture_comparison}, including GAT and other graph- or image-based baseline models, were evaluated under the same experimental protocol and provided with the same auxiliary input information, including age, education level, and NPT features. As shown in Table~\ref{tab:gnn_architecture_comparison}, both advanced GNN variants achieve competitive performance. Polynormer attains the second-highest accuracy (0.884; $p=0.002$), F1 score (0.684; $p=0.010$), recall (0.603; $p=0.027$), and balanced accuracy (0.785; $p=0.003$), while EGNN exhibits the second-highest AUROC (0.924; $p=0.168$), AUPRC (0.874; $p=0.858$), and precision (0.899; $p=0.121$). Nevertheless, the attention-based GAT achieves the strongest and most balanced performance overall.

Interestingly, unlike earlier observations, Polynormer no longer exhibits near-perfect precision but instead demonstrates more balanced behavior with precision of 0.855 and recall of 0.603. EGNN, meanwhile, achieves ranking performance close to GAT, particularly in AUROC and AUPRC, but lower recall limits its overall classification performance. In contrast, GAT maintains stronger balance between sensitivity and specificity, leading to superior overall F1 and balanced accuracy. These findings suggest that increasing architectural complexity does not necessarily translate to more reliable classification performance in small-cohort clinical settings.

Importantly, our objective is not to exhaustively benchmark GNN architectures, but rather to adopt a model that is sufficiently expressive while remaining amenable to interpretation. Using a relatively lightweight attention-based GNN enables meaningful SHAP-based analyses, allowing us to disentangle contributions from multiple modalities and identify discriminative structural patterns such as graphlets associated with AD-related drawing abnormalities. In this context, the chosen architecture provides an effective balance between predictive performance and interpretability, supporting the broader goal of transparent and clinically meaningful model analysis.

\subsection{Comparison with Pixel-Based Baselines}
To assess the benefit of graph-based modeling over conventional image-based approaches, we compared our method against two pixel-based convolutional baselines commonly used in digital drawing analysis: a lightweight CNN and ResNet-18 \cite{he2016deep}. The CNN consisted of two convolutional layers with 16 and 32 output channels, each followed by ReLU activation and max-pooling operations for hierarchical feature extraction and spatial downsampling. The resulting feature maps were flattened and passed through a fully connected layer with 128 hidden units to generate compact image representations for classification. Both models were trained and evaluated on rasterized cube drawings using the same experimental protocol and auxiliary input features to ensure a fair comparison. As summarized in Table~\ref{tab:gnn_architecture_comparison}, both pixel-based models perform substantially worse than graph-based methods across all evaluation metrics. The performance gap is particularly evident in F1 score and AUPRC. The CNN achieves an F1 score of only 0.320 and AUPRC of 0.481, while ResNet-18 reaches an F1 score of 0.333 and AUPRC of 0.497. In comparison, all graph-based models achieve F1 scores above 0.65 and AUPRC values above 0.84. Pixel-based approaches also exhibit markedly lower recall and balanced accuracy, suggesting reduced sensitivity to AD-related drawing abnormalities.

These results suggest that raw pixel representations fail to capture the subtle structural and relational characteristics underlying impaired cube-copying behavior in Alzheimer's disease. In contrast, graph representations explicitly encode visuospatial relationships such as corner connectivity and edge continuity, preserving higher-level structural information relevant to cognitive impairment. Even advanced image-based architectures such as ResNet-18 show only marginal improvements over the lightweight CNN baseline, indicating that increased representational capacity alone may not compensate for the lack of explicit structural encoding. This structured abstraction not only improves predictive performance in limited-sample settings but also facilitates downstream interpretability analyses, reinforcing the value of graph-based representations for robust and explainable AD classification.

\section{Discussion}
\subsection{Summary of Results}
Our findings demonstrate that cube-copying behavior, when represented through graph-structured modeling, provides meaningful and quantifiable biomarkers for Alzheimer’s disease. Even in the unimodal setting, cube graph representations achieve strong discriminative performance, indicating that visuospatial construction errors manifest as systematic structural disruptions that can be captured through graph topology and geometry. This observation supports longstanding clinical evidence that cube-copying tasks are sensitive to visuospatial and executive dysfunction and further suggests that such impairments can be translated into machine-interpretable representations.

The integration of demographic and neuropsychological information further improves performance by providing complementary contextual signals. In particular, age and education substantially improve sensitivity and balanced classification performance, as reflected by gains in recall and balanced accuracy in multimodal settings. NPT information contributes additional complementary information that further improves discriminative performance, particularly AUROC and AUPRC. Interestingly, SHAP analysis indicates that NPT variables do not emerge as dominant standalone predictors, suggesting that their contribution may arise primarily through interactions with structural and demographic information. These observations highlight the value of multimodal integration, where different modalities contribute distinct yet complementary aspects of disease-related information.

Importantly, multimodal fusion improves not only predictive performance but also interpretability. SHAP-based analyses reveal that specific graphlet motifs and geometric disruptions, such as reduced corner integrity and weakened edge continuity, emerge as highly influential predictors. In particular, graphlet 6 and graphlet 4 correspond to characteristic cube structures that frequently deteriorate in AD drawings. These findings align closely with clinical descriptions of incomplete, disorganized, and distorted drawing patterns observed in individuals with Alzheimer's disease, reinforcing the clinical plausibility of the model's decision-making process.

\subsection{Clinical Implications}
From a practical perspective, the proposed framework highlights the potential of cube-copying tasks as scalable, low-cost, and non-invasive screening tools. By relying on simple drawings rather than expensive imaging procedures or invasive biomarkers, the approach is well suited for deployment in community settings, primary care environments, or large-scale cognitive screening programs. Moreover, graph-based representations preserve higher-level structural relationships and support interpretable analysis in relatively small datasets, a common challenge in clinical studies.

\subsection{Limitations}
Several limitations should be acknowledged. First, the current task focuses on binary classification between Alzheimer's disease and cognitively normal controls, which does not fully capture the continuum of cognitive decline. Second, although the cohort is clinically curated, its modest sample size limits generalizability and motivate future validation using larger and more diverse populations. Third, the present framework relies primarily on static structural properties of drawings and does not exploit temporal characteristics such as stroke order, speed, pauses, or hesitation patterns, which may provide additional diagnostic information.

\subsection{Future Directions}
Future work will address these limitations by extending the framework to multi-class settings involving cognitively normal individuals, mild cognitive impairment, and Alzheimer's disease. Such extensions are particularly important because MCI represents a prodromal stage of AD where earlier intervention may be most beneficial. Incorporating temporal drawing dynamics and longitudinal observations may further improve sensitivity to subtle cognitive changes over time. Together, these directions position graph-based analysis of drawing behavior as a promising avenue toward early, interpretable, and accessible assessment of neurodegenerative disease.

\section{Funding}
K.Y. is supported in part by the National Research Foundation of Korea (NRF) grant (No. RS-2024-00337092), the Institute of Information \& communications Technology Planning \& Evaluation (IITP) grants (No. RS-2020-II201373, Artificial Intelligence Graduate School Program; No. IITP-(2025)-RS-2023-00253914, Artificial Intelligence Semiconductor Support Program (Hanyang University)) funded by the Korean government (MSIT).


\section{Data Availability}
The datasets generated and/or analyzed during the current study are not publicly available due to ethical restrictions imposed by the institutional review board and the sensitive nature of the clinical and behavioral data. However, they may be made available from the corresponding author upon reasonable request and subject to institutional approval.

\clearpage
\bibliographystyle{named}
\bibliography{references}

\clearpage
\appendix
\onecolumn
\section{Supplementary Material}\label{sec:SI}
To further assess whether the observed performance differences were statistically supported, we conducted pairwise comparisons between the best-performing model and each baseline model under the same experimental setting. Statistical significance was evaluated using two complementary tests: \textit{t}-tests, which compare differences in mean performance between models, and Kolmogorov--Smirnov tests, which compare the overall distributions of performance scores without assuming that the difference is limited to the mean. Both tests were applied to the results from 10 repetitions of 5-fold cross-validation for each evaluation metric. The resulting \textit{p}-values are reported in the following tables, where smaller \textit{p}-values indicate stronger statistical evidence of a performance difference between the best-performing model and the corresponding baseline.

\begin{table}[h]
\centering
\caption{Statistical comparison of model performance using \textit{t}-tests for the results reported in Table~\ref{tab:pred_multimodal}. Each reported \textit{p}-value corresponds to a comparison between the best-performing model, \textit{Cube graph + Age + Edu + NPT}, and the corresponding baseline model under the same experimental setting. Smaller \textit{p}-values indicate stronger statistical evidence of a performance difference.}
\label{tab:ttest_pred_multimodal}
\resizebox{\textwidth}{!}{%
\begin{tabular}{lccccccc}
\toprule
Model & Accuracy & F1 & AUC & AUPRC & Precision & Recall & Balanced Acc. \\
\midrule
Cube graph + Age + Edu + NPT & -- & -- & -- & -- & -- & -- & -- \\
Cube graph + Age + Edu       & \(p = .464\) & \(p = .817\) & \(p < .001\) & \(p = .026\) & \(p < .001\) & \(p = .127\) & \(p = .409\) \\
Cube graph + Age + NPT       & \(p < .001\) & \(p = .001\) & \(p = .185\) & \(p = .418\) & \(p < .001\) & \(p < .001\) & \(p < .001\) \\
Cube graph + Edu + NPT       & \(p < .001\) & \(p < .001\) & \(p = .031\) & \(p = .010\) & \(p < .001\) & \(p < .001\) & \(p < .001\) \\
Cube graph                   & \(p = .001\) & \(p < .001\) & \(p = .873\) & \(p = .562\) & \(p = .007\) & \(p < .001\) & \(p < .001\) \\
Age + Edu + NPT              & \(p < .001\) & \(p < .001\) & \(p < .001\) & \(p < .001\) & \(p < .001\) & \(p < .001\) & \(p < .001\) \\
\bottomrule
\end{tabular}}
\end{table}

\begin{table}[h]
\centering
\caption{Statistical comparison of model performance using \textit{t}-tests for the results reported in Table~\ref{tab:gnn_architecture_comparison}. Each reported \textit{p}-value corresponds to a comparison between the best-performing model, \textit{GAT}, and the corresponding baseline model under the same experimental setting.}
\label{tab:ttest_gnn_architecture_comparison}
\resizebox{0.8\textwidth}{!}{%
\begin{tabular}{lccccccc}
\toprule
Model & Accuracy & F1 & AUC & AUPRC & Precision & Recall & Balanced Acc. \\
\midrule
GAT        & -- & -- & -- & -- & -- & -- & -- \\
Polynormer & \(p = .002\) & \(p = .010\) & \(p = .004\) & \(p = .038\) & \(p = .030\) & \(p = .027\) & \(p = .003\) \\
EGNN       & \(p = .005\) & \(p = .003\) & \(p = .168\) & \(p = .858\) & \(p = .121\) & \(p = .004\) & \(p = .003\) \\
CNN        & \(p < .001\) & \(p < .001\) & \(p < .001\) & \(p < .001\) & \(p < .001\) & \(p < .001\) & \(p < .001\) \\
ResNet-18  & \(p < .001\) & \(p < .001\) & \(p < .001\) & \(p < .001\) & \(p < .001\) & \(p < .001\) & \(p < .001\) \\
\bottomrule
\end{tabular}}
\end{table}

\begin{table}[h]
\centering
\caption{Statistical comparison of model performance using Kolmogorov--Smirnov tests for the results reported in Table~\ref{tab:pred_multimodal}. Each reported \textit{p}-value corresponds to a comparison between the best-performing model, \textit{Cube graph + Age + Edu + NPT}, and the corresponding baseline model under the same experimental setting.}
\label{tab:ks_pred_multimodal}
\resizebox{\textwidth}{!}{%
\begin{tabular}{lccccccc}
\toprule
Model & Accuracy & F1 & AUC & AUPRC & Precision & Recall & Balanced Acc. \\
\midrule
Cube graph + Age + Edu + NPT & -- & -- & -- & -- & -- & -- & -- \\
Cube graph + Age + Edu       & \(p = .293\) & \(p = .715\) & \(p = .007\) & \(p = .040\) & \(p = .018\) & \(p = .164\) & \(p = .164\) \\
Cube graph + Age + NPT       & \(p = .055\) & \(p = .044\) & \(p = .263\) & \(p = .573\) & \(p = .044\) & \(p = .029\) & \(p = .007\) \\
Cube graph + Edu + NPT       & \(p = .015\) & \(p = .029\) & \(p = .164\) & \(p = .029\) & \(p = .003\) & \(p = .015\) & \(p = .003\) \\
Cube graph                   & \(p = .007\) & \(p < .001\) & \(p = .350\) & \(p = .164\) & \(p = .023\) & \(p < .001\) & \(p < .001\) \\
Age + Edu + NPT              & \(p < .001\) & \(p < .001\) & \(p = .002\) & \(p < .001\) & \(p < .001\) & \(p < .001\) & \(p < .001\) \\
\bottomrule
\end{tabular}}
\end{table}

\begin{table}[h]
\centering
\caption{Statistical comparison of model performance using Kolmogorov--Smirnov tests for the results reported in Table~\ref{tab:gnn_architecture_comparison}. Each reported \textit{p}-value corresponds to a comparison between the best-performing model, \textit{GAT}, and the corresponding baseline model under the same experimental setting.}
\label{tab:ks_gnn_architecture_comparison}
\resizebox{0.8\textwidth}{!}{%
\begin{tabular}{lccccccc}
\toprule
Model & Accuracy & F1 & AUC & AUPRC & Precision & Recall & Balanced Acc. \\
\midrule
GAT        & -- & -- & -- & -- & -- & -- & -- \\
Polynormer & \(p = .003\) & \(p = .023\) & \(p = .023\) & \(p = .003\) & \(p = .055\) & \(p = .055\) & \(p = .003\) \\
EGNN       & \(p = .002\) & \(p = .007\) & \(p = .350\) & \(p = .350\) & \(p = .350\) & \(p = .007\) & \(p = .002\) \\
CNN        & \(p < .001\) & \(p < .001\) & \(p < .001\) & \(p < .001\) & \(p < .001\) & \(p < .001\) & \(p < .001\) \\
ResNet-18  & \(p < .001\) & \(p < .001\) & \(p < .001\) & \(p < .001\) & \(p < .001\) & \(p < .001\) & \(p < .001\) \\
\bottomrule
\end{tabular}}
\end{table}

\end{document}